\documentclass[sigconf]{aamas} 

\usepackage{balance} 

\usepackage{hyperref}
\usepackage{url}

\usepackage{graphicx}
\usepackage{algorithm}
\usepackage{algpseudocode}
\usepackage{enumitem}
\usepackage{caption}
\usepackage{tikz} 
\usepackage{xcolor}
\usepackage{subcaption}
\usepackage{pgfplots}
\usepackage{adjustbox}

\renewcommand\footnotetextcopyrightpermission[1]{}

\definecolor{tab_option1}{HTML}{1f77b4}
\definecolor{tab_option2}{HTML}{ff7f0e}
\definecolor{tab_option3}{HTML}{2ca02c}
\definecolor{tab_option4}{HTML}{9467bd}

\definecolor{reach_moc2her2}{HTML}{1f77b4}
\definecolor{reach_ioc2her2}{HTML}{4fa8db}
\definecolor{reach_moc2her4}{HTML}{ff7f0e}
\definecolor{reach_ioc2her4}{HTML}{ffb366}
\definecolor{reach_moc2her8}{HTML}{2ca02c}
\definecolor{reach_ioc2her8}{HTML}{6fdc6f}
\definecolor{reach_moc2}{HTML}{9467bd}
\definecolor{reach_ioc2}{HTML}{b699d9}
\definecolor{reach_moc4}{HTML}{8c564b}
\definecolor{reach_ioc4}{HTML}{b98c80}
\definecolor{reach_moc8}{HTML}{e377c2}
\definecolor{reach_ioc8}{HTML}{f2a9db}

\definecolor{moc2her}{RGB}{0,114,178}    
\definecolor{ioc2her}{RGB}{255,127,14}  
\definecolor{mocher}{RGB}{148,103,189}  
\definecolor{iocher}{RGB}{140,86,75}    
\definecolor{moc}{RGB}{44,160,44}       
\definecolor{ioc}{RGB}{228,65,28}       

\setcopyright{none}
\acmConference{}{}{}
\copyrightyear{}
\acmYear{}
\acmDOI{}
\acmPrice{}
\acmISBN{}

\title{Enabling Option Learning in Sparse Rewards \\with Hindsight Experience Replay}
\thanks{This paper has been accepted as an extended abstract at the 2026 International Conference on Autonomous Agents and Multiagent Systems (AAMAS)~\citep{romio:2026}}

\author{Gabriel Romio}
\email{gromio@edu.unisinos.br}
\affiliation{%
  \institution{Universidade do Vale do Rio dos Sinos}
  \city{São Leopoldo}
  \state{RS}
  \country{Brazil}
}

\author{Mateus Begnini Melchiades}
\email{mateusbme@edu.unisinos.br}
\affiliation{%
  \institution{Universidade do Vale do Rio dos Sinos}
  \city{São Leopoldo}
  \state{RS}
  \country{Brazil}
}

\author{Bruno Castro da Silva}
\email{bsilva@cs.umass.edu}
\affiliation{%
  \institution{University of Massachusetts, Amherst}
  \city{Amherst}
  \state{MA}
  \country{USA}
}

\author{Gabriel de Oliveira Ramos}
\email{gdoramos@unisinos.br}
\affiliation{%
  \institution{Universidade do Vale do Rio dos Sinos}
  \city{São Leopoldo}
  \state{RS}
  \country{Brazil}
}

\begin{abstract}
Hierarchical Reinforcement Learning (HRL) frameworks like Option-Critic (OC) and Multi-updates Option Critic (MOC) have introduced significant advancements in learning reusable options. However, these methods underperform in multi-goal environments with sparse rewards, where actions must be linked to temporally distant outcomes. To address this limitation, we first propose MOC-HER, which integrates the Hindsight Experience Replay (HER) mechanism into the MOC framework. By relabeling goals from achieved outcomes, MOC-HER can solve sparse reward environments that are intractable for the original MOC. However, this approach is insufficient for object manipulation tasks, where the reward depends on the object reaching the goal rather than on the agent’s direct interaction. This makes it extremely difficult for HRL agents to discover how to interact with these objects. To overcome this issue, we introduce Dual Objectives Hindsight Experience Replay (2HER), a novel extension that creates two sets of virtual goals. In addition to relabeling goals based on the object's final state (standard HER), 2HER also generates goals from the agent's effector positions, rewarding the agent for both interacting with the object and completing the task. 
Experimental results in robotic manipulation environments show that MOC-2HER achieves success rates of up to 90\%, compared to less than 11\% for both MOC and MOC-HER.
These results highlight the effectiveness of our dual objective relabeling strategy in sparse reward, multi-goal tasks.
\end{abstract}


\newcommand{\BibTeX}{\rm B\kern-.05em{\sc i\kern-.025em b}\kern-.08em\TeX}

\begin{document}


\pagestyle{fancy}
\fancyhead{}


\maketitle 


\section{Introduction}

The concept of Hierarchical Reinforcement Learning (HRL)~\citep{dayan1993feudal, Sutton:1999, klissarov2025discovering} aims to structure complex tasks into more manageable sub-tasks. In this context, the Options framework~\citep{Sutton:1999} enables the reuse of learned skills between different problems, thus accelerating the training process. Options are closed-loop policies capable of executing a complete action over an extended period of time~\citep{Sutton:2020}. Examples of options include tasks like picking up an object, or traveling between two cities, while primitive actions are fundamental operations, such as muscle twitches or applying electrical current to a motor~\citep{nachum2019does}.

The Option-Critic (OC) framework~\citep{Bacon_Harb_Precup_2017} initiated automatic end-to-end option learning, representing a foundational contribution that has subsequently guided advances in enhancing efficiency and accelerating training. Subsequently, the Multi-updates Option Critic (MOC) framework~\citep{NEURIPS2021_24cceab7} enabled simultaneous updates of multiple options, significantly improving performance in high-complexity environments and preventing the degeneration of solutions. Despite this, complex tasks with sparse rewards remain a significant challenge in the context of HRL, since discovering and learning options without well-defined rewards is particularly difficult~\citep{Pateria:2021, klissarov2025discovering}. The Hindsight Experience Replay (HER) algorithm~\citep{NIPS2017_453fadbd}, by recalculating rewards from failed trajectories through a replay buffer, introduced significant improvements in handling sparse rewards in multi-goal environments. 

Following this concept, we extend MOC with the HER mechanism to handle option learning in sparse reward scenarios. This method aims to improve or enable the discovery of options in these environments in a straightforward approach. To achieve this, we designed a replay buffer to store information about each state transition from the original MOC, including the active option. These states are then reevaluated with a new goal, selected by one of the states reached by the algorithm. The trajectory is reassessed as if the objective had been to reach that state from the beginning, thus the experience can be utilized even if the trajectory did not achieve its original goal. Each state stored in the HER buffer has its reward value recalculated based on the new goal. The HER buffer is subsequently integrated into the MOC training buffer, becoming part of the evaluation and learning phases of the algorithm. We refer to this enhanced framework as \textit{Multi-Updates Option Critic with Hindsight Experience Replay} (MOC-HER).

However, complex object manipulation remains challenging for HRL, as agents must interact with objects to achieve a goal that can change with each episode, an intrinsic feature of multi-goal environments~\citep{plappert2018multigoalreinforcementlearningchallenging}. In these environments, the reward signal is often tied only to the final state of the object (e.g., the object reaching its target position) and not to the actions of the agent (e.g., the gripper of an industrial robot). This means the agent is not directly rewarded for interacting with the object, leading to a significant increase in learning complexity. While the original HER addresses these object manipulation tasks in standard RL~\citep{NIPS2017_453fadbd}, it is insufficient for HRL, where learning interaction dynamics and multiple policies simultaneously is required. Without a well-defined reward signal, this learning becomes unguided, hindering the convergence of the hierarchical components.

To overcome this limitation, we further propose \textit{Dual Objectives Hindsight Experience Replay} (2HER), an extension that introduces a second objective to the hindsight mechanism, explicitly designed to encourage interaction between the agent and the object. In addition to generating virtual goals based on the positions reached by the object (conventional HER), our methodology also creates a second set of goals from the subsequent positions reached by the agent's effector. In other words, the agent is retrospectively rewarded not only for taking the object to a hypothetical destination but also for moving the gripper to a position where the object once was. The final reward signal used for learning combines both objectives, simultaneously encouraging the exploration of interaction and the completion of the task.

We conducted extensive experiments with MOC-HER and MOC-2HER in the Fetch Robotics environments~\citep{plappert2018multigoalreinforcementlearningchallenging}, where we varied the number of options and performed comparative studies involving several baseline algorithms. The results demonstrate that both HER and 2HER can solve sparse reward environments that are challenging for standard HRL approaches. Specifically, in challenging object manipulation tasks, our 2HER-augmented method achieved a success rate of up to 90\%, a significant improvement over the 11\% achieved by the original algorithm. To the best of our knowledge, this is the first approach to tackle sparse rewards using the MOC framework.

Our contributions can be summarized as follows:

\begin{itemize}[nosep]
    \item We integrate standard HER into the MOC framework, resulting in an algorithm capable of solving simple sparse reward environments without object manipulation.    
    \item We devise 2HER, a novel extension that improves the ability of HRL algorithms to handle object manipulation tasks in sparse reward settings. We further combine 2HER with MOC to address these complex environments while maintaining high-quality option generation.
    \item We conduct extensive experiments that show success rates increasing from 0, with MOC, to 100\%, with MOC-HER, in sparse reward environments and from 11\%, with MOC, to 90\%, with MOC-2HER, on object manipulation tasks.
\end{itemize}

\section{Background}



\textbf{Hierarchical Reinforcement Learning} utilizes temporal abstractions~\citep{precup2000temporal}, called options, to execute temporally extended actions~\citep{Sutton:1999}. Each option \(o\) is a tuple $(I_o, \pi_o, \beta_o)$, defined by an intra-option policy \( \pi_o : S \times A \to [0, 1] \), a termination condition \( \beta_o : S \to [0, 1] \), and an initiation set \( I_o \subseteq S \). Once an option is initiated, actions are chosen according to \( \pi_o \) until termination, as sampled by \(\beta(s,o)\). Based on the policy gradient theorem~\citep{sutton2000policy}, the Option-Critic (OC) architecture~\citep{Bacon_Harb_Precup_2017} enables the end-to-end learning of these options. However, a limitation is that it updates only the single option active at each timestep, making learning particularly challenging in sparse reward tasks.

\textbf{Multi-updates Option Critic} algorithm~\citep{NEURIPS2021_24cceab7} extends the intra-option learning framework by enabling simultaneous, off-policy updates for all relevant options in a given state, rather than only for the one being executed. This approach promotes the development of temporally extended options and prevents options from degenerating, where some options are chosen much more frequently than others.
The MOC framework uses a policy over options \(\mu(o|s)\) to select an option $o$ in a state \(s\). The subsequent actions \(a\) are determined by the corresponding intra-option policy \(\pi_{\zeta}(a|s, o)\), parameterized by $\zeta$. The learning process relies on the value function $Q_O(s, o)$, which operates as the algorithm's critic by assigning expected return estimates to the action executed under a given option.
To update all relevant options simultaneously, MOC employs an off-policy approach that leverages intra-option policy gradients (the actor). This gradient is represented by \(\log \partial \pi_{\zeta}\), where the policy \(\pi_{\zeta}\) is updated using the past experience. The update process uses the occupancy distribution \(p_{\mu,\beta}\), which reflects how frequently each option is active in a given state. The updates are therefore weighted according to each option's relevance in the observed states.

\textbf{Multi-Goal Reinforcement Learning} involves a set of continuous control tasks with sparse rewards~\citep{plappert2018multigoalreinforcementlearningchallenging}. All of these environments feature a goal that changes after each episode and encoded in the observation space, as formalized by ~\citet{pmlr-v37-schaul15}, defining a \textit{goal-based} environment. Thus, the observation space contains, in addition to the current environment states, two new parameters: the \textit{desired goal}, which represents the target that the agent needs to reach, and \textit{achieved goal}, which represents the position the agent has reached at the current timestep. In tasks involving object manipulation, such as a robotic arm learning to move objects in the scene, this value corresponds to the object's current position. Standard RL algorithms struggle in such tasks, as the highly sparse reward function provides insufficient feedback for policy optimization.

\textbf{Hindsight Experience Replay} algorithm~\citep{NIPS2017_453fadbd} addresses this challenge by reinterpreting failed trajectories, redefining the desired goal. After completing an episode \(s_0, s_1, \ldots, s_T\), each transition \( s_t \to s_{t+1} \) is stored in a replay buffer. To enhance the learning data, alternative goals are introduced for these transitions. These new goals are sampled through a strategy $\mathbb{S}$, such as the final state achieved (\texttt{final}), a random future state on the same trajectory (\texttt{future}) or a random state from the entire episode (\texttt{episode}). The rewards for these transitions are recomputed as if the new goal had always been the target. This approach enables the agent to gain insights about how to reach this state, which can be leveraged by any off-policy RL algorithm.

\section{Related work}

Traditional HRL methods focus on learning options end-to-end. The Option-Critic (OC)~\citep{Bacon_Harb_Precup_2017} introduced a framework for option learning with policy gradient. Asynchronous Advantage Option-Critic (A2OC)~\citep{harb2017waitingoptionlearning} implemented the \textit{deliberation cost} to improve and enhance the stability to the use of options, while Proximal Policy Option-Critic (PPOC)~\citep{klissarov2017learningsoptionsendtoendcontinuous} was among the first options-based approaches to demonstrate strong performance in continuous control tasks. The Interest Option-Critic (IOC)~\citep{Khetarpal_Klissarov_Chevalier-Boisvert_Bacon_Precup_2020} refined option specialization through interest functions, allowing each option to focus on specific regions of the state space. More recently, Dynamic Option Creation~\citep{Melchiades:2025} proposes an approach for dynamically creating new options during the training process.

However, despite their advances, these frameworks face difficulties in environments with sparse rewards, a challenge highlighted by~\citet{Pateria:2021}. Common strategies to mitigate this, such as transfer learning~\citep{taylor2011introduction}, techniques that encourage exploration~\citep{pathak2017curiosity, eysenbach2018diversity, NEURIPS2022_266c0f19} and sub-goal discovery~\citep{mcgovern2001automatic, kulkarni2016hierarchical}, often introduce their own limitations. 
These include reliance on prior data or domain-specific rules and the risk of inefficient or unguided exploration, which limits the agent’s autonomy and its ability to tackle novel or more complex tasks.

In the context of robotics or multi-goal tasks, and as an alternative to options, \textit{skills}~\citep{levine2015learning, lee2021pebble, wang2022skill} enable an agent to learn general behaviors from interactions with the environment, allowing effective reuse and transfer. However, unlike options, skills are typically not learned end-to-end and are often driven by intrinsic motivation or human feedback.

Furthermore, goal relabeling techniques incorporated into HRL have been explored previously. ~\citet{nachum2018data} extended off-policy correction to handle tasks with sparse rewards. They proposed reusing past experience by relabeling goals at a higher level to guide a lower-level policy. While this approach achieved strong performance on locomotion tasks, it exhibited training instability. ~\citet{levy2019learningmultilevelhierarchieshindsight} mitigated this instability by training each hierarchical level separately, thereby improving upon prior results, but their method does not address object manipulation tasks.

\section{Method}

In this work, we incorporated a HER buffer into the MOC framework to improve its performance in multi-goal environments, particularly those characterized by sparse rewards. The integration preserves the fundamental architecture of MOC, thus retaining its inherent advantages while addressing the challenges of reward sparsity. Our approach, detailed in Algorithm 1 (Section~\ref{sec:Algorithm}), involves recording data for each state transition into a replay buffer throughout the agent's interaction with the environment. This data comprises not only the actual state of the environment but also the obtained rewards and the selected options. Virtual goals are then sampled from states reached by the agent according to the goal-sampling strategy $\mathbb{S}$, and the reward for each stored transition is recomputed with respect to the sampled goal. This enables unsuccessful trajectories to be repurposed for learning, treating them as if the intended goal had been achieved.

When the environment involves object manipulation, a secondary objective is incorporated into the hindsight process.
This is achieved by augmenting the standard HER mechanism. 
In addition to relabeling transitions with a future object position (the primary goal), a supplementary set of transitions is created by using a future agent position as a secondary goal. This extended retrospective mechanism, denoted as Dual Objectives Hindsight Experience Replay (2HER), provides the essential experiences required for learning the object manipulation objective.
The reward is then redefined as a combination of two binary components: the original reward $r_{goal}$, associated with task completion, and an additional reward $r_{object}$, designed to incentivize interaction between the agent and the object. These components are weighted by a coefficient $C_r \in [0,1]$ that determines their relative contribution to the learning.

During evaluation and learning phases, the HER buffer is merged with the buffer containing the original trajectories. 
A detailed explanation of HER implementation in MOC is provided in Section~\ref{sec:MOC_HER}.

\subsection{Algorithm} \label{sec:Algorithm}

A description of the implementation is provided in Algorithm~\ref{alg:MOC_HER_algorithm}, which extends the original MOC algorithm~\citep{NEURIPS2021_24cceab7} with our 2HER approach. The algorithm outlines a simplified version of the process, focusing on a single episode to provide a clearer understanding of the overall approach. Our modifications to the original algorithm begin at line 1 and 4, where we introduce the initialization of the strategy $\mathbb{S}$ for sampling new goals for replay at each state along the trajectory $(s_0, \ldots, s_T)$, the reward function $r(s_t,a_t,g)$, and the HER buffer $R$. Additionally, component 1 of Algorithm~\ref{alg:MOC_HER_algorithm} (lines 15-18) stores transitions to new states in the replay buffer. This process includes the selection of the new goal,
the calculation of the new reward based on this goal, and the storage of the transition with the updated values in the HER buffer (more details in Section~\ref{sec:MOC_HER}). 

When object interaction is required, which is automatically detected through an intrinsic property of the environment, the 2HER component is introduced as a secondary objective that applies hindsight analysis to the agent–object positioning (lines 19–29).
At line 22, the transition reward is recalculated using the reward coefficient $C_r$. This coefficient determines the relative contribution to the final reward of two components: the original task reward ($r_{goal}$) and the object-interaction objective reward ($r_{object}$). In scenarios where object manipulation is not required, the algorithm defaults to MOC-HER, disregarding the additional 2HER component.

This procedure is applied to each state transition within the episode. For the learning phase, the evaluation and improvement steps (lines 30 and 36) are performed on mini-batches of transitions. These mini-batches are sampled from a buffer that merges the original experiences with the relabeled hindsight transitions from the buffer \( R \).

\setlength{\textfloatsep}{8pt plus 1.0pt minus 2.0pt}

\begin{algorithm}[t]
\caption{Multi-updates Option Critic with Dual Objectives Hindsight Experience Replay (MOC-2HER)}
\label{alg:MOC_HER_algorithm}
\footnotesize 
\begin{algorithmic}[1]
\State \textcolor{red}{\textbf{input:} $\mathbb{S}$ (Strategy for sampling goals for replay, e.g. $\mathbb{S}{(s_0, \ldots, s_T)}$); $r(s_t,a_t,g)$ (Function to recalculate the rewards)}
\State Set $s \gets s_0$
\State Choose $o$ at $s$ according to $\mu_z(\cdot | s)$
\State \textcolor{red}{Initialize empty HER buffer $R$}
\Repeat
    \State Choose $a$ according to $\pi_\zeta(a|s, o)$
    \State Take action $a$ in $s$, observe $s_{t+1}$, $r$
    \State Sample termination from $\beta_\nu(s_{t+1}, o)$
    \If{$o$ terminates in $s_{t+1}$} 
        \State Sample $o_{t+1}$ according to $\mu_z(\cdot | s_{t+1})$
    \Else
        \State $o_{t+1} = o$
    \EndIf
    \State Define previous option $\bar{o} = o$

    \State \textcolor{red}{\textbf{1. Hindsight Experience Replay:}{\Comment{start of HER implementation}}}
    \State \textcolor{red}{Obtain the transition $(s_t, o_t, a_t, r_t, s_{t+1})$ from the latest step}
    \algrenewcommand\algorithmicfor{\textcolor{red}{\textbf{for}}}
    \algrenewcommand\algorithmicdo{\textcolor{red}{\textbf{do}}}
    \algrenewcommand\algorithmicif{\textcolor{red}{\textbf{if}}}
    \algrenewcommand\algorithmicthen{\textcolor{red}{\textbf{then}}}
    \algrenewcommand\algorithmicelse{\textcolor{red}{\textbf{else}}}
    \algrenewcommand\algorithmicend{\textcolor{red}{\textbf{end}}} 
    \State \textcolor{red}{$g' \gets \mathbb{S_{\text{goal}}}{\text{(current transition)}}$}
    \State \textcolor{red}{Recalculate $r'_{\text{goal}}$ based on the function $r(s_t,a_t,g)$ for the new goal $g'$}        
    \If {\textcolor{red}{has\_object}} \textcolor{red}{\Comment{start of 2HER implementation}}
        \State \textcolor{red}{$s'_{\text{obj\_pos}} \gets \mathbb{S}_{\text{agent\_pos}}$ (current transition) \Comment{Samples an agent effector position to use as the new object position}}
        \State \textcolor{red}{Calculate $r'_{\text{obj}}$ with the function $r(s_t,a_t,g)$ for the distance agent\_object}
        \State \textcolor{red}{$r' \gets C_r r_{\text{obj}} + (1-C_r) r_{\text{goal}}$}
    \Else
        \State \textcolor{red}{$s'_{\text{obj\_pos}} \gets s_{\text{obj\_pos}}$}
        \State \textcolor{red}{$r' \gets r_{\text{goal}}$}
    \EndIf 
    \State \textcolor{red}{$\bar{s}_{t} \gets s_{t} \;\|\; g'$ \Comment{$\;\|\;$ denotes concatenation}}
    \State \textcolor{red}{$\bar{s}_{t+1} \gets s_{t+1} \;\|\; s'_{\text{obj\_pos}} \;\|\; g'$}
    \State \textcolor{red}{Store transition $(\bar{s}_{t}, o_{t}, a_{t}, r', \bar{s}_{t+1})$ in HER buffer $R$}
    
    \algrenewcommand\algorithmicfor{for}
    \algrenewcommand\algorithmicdo{do}
    \algrenewcommand\algorithmicif{if}
    \algrenewcommand\algorithmicthen{then}
    \algrenewcommand\algorithmicelse{else}
    \algrenewcommand\algorithmicend{end}
    
    \State \textbf{2. Evaluation step:}
    \For{each option $\tilde{o}$ in the option set $O$}
        \State $\delta \gets \mathbb{E}[U^\rho | s, \tilde{o}] - Q_\theta(s, \tilde{o})$ \text{ where } $U^\rho$ is an importance sampling weighted target
        \State $\theta \gets \theta + p_{\mu, \beta}(\tilde{o} | s, \bar{o}) \alpha \delta \phi(s, \tilde{o})$
    \EndFor
    \State \textcolor{red}{Repeat this step for mini-batches of HER transitions $(\bar{s}_{t}, o_{t}, a_{t}, r', \bar{s}_{t+1})$}
    
    \State \textbf{3. Improvement step:}
    \For{each option $\tilde{o}$ in the option set $O$}
        \State $\zeta \gets \zeta + p_{\mu, \beta}(\tilde{o} | s, \bar{o}) \alpha_\zeta \frac{\partial \log \pi_\zeta(a|s,o)}{\partial \zeta} Q_\theta(s, o, a)$
    \EndFor
    \State \textcolor{red}{Repeat this step for mini-batches of HER transitions $(\bar{s}_{t}, o_{t}, a_{t}, r', \bar{s}_{t+1})$}
    \State $\nu \gets \nu - \alpha_\nu \frac{\partial \beta_\nu(s_{t+1}, o)}{\partial \nu} (Q_\theta(s_{t+1}, o) - V_\theta(s_{t+1}))$
    \State $z \gets z + \alpha_z \beta_\nu(s_{t+1}, o) \frac{\partial \mu_z(o_{t+1} | s_{t+1})}{\partial z} Q_\theta(s_{t+1}, o_{t+1})$
    \State $s \gets s_{t+1}$
\Until{$s'$ is a terminal state}
\end{algorithmic}
\end{algorithm}

\subsection{Implementing HER into MOC} \label{sec:MOC_HER}

In this section, we provide detailed information about the HER implementation.
As previously mentioned, to improve MOC performance in multi-goal RL environments, we devise a replay buffer that is incremented after each timestep. 
This buffer stores key information about the decisions and current state, which will be used for over-option and intra-option learning. This stored data includes the action performed by the agent, the option number that executed the action, the observed state, the resulting state, and the reward received following the action.

After each episode concludes, we process the just-completed trajectory in the replay buffer with HER.
To define the new goals we use the strategy $\texttt{future}$, chosen for its strong performance in the original HER experiments~\citep{NIPS2017_453fadbd}.
According to this strategy, for each transition \((s_t,o_t,a_t,r_t,s_{t+1})\) in the original trajectory, we randomly sample a set of $k$ additional goals \(\{g_1',...,g_k'\}\) from future states of the episode.
The reward is then reevaluated for each transition based on these new goals, utilizing the reward function inherent to the selected environment.
For the sparse reward setting considered in this work, the value is binary: the reward is $0$ if the achieved goal matches the desired goal in a given transition, and $-1$ otherwise.



We introduce the 2HER extension to improving learning in object manipulation scenarios with sparse reward. In such tasks, the agent typically receives a reward signal only when an object reaches its target position, making it difficult to learn the prerequisite interaction and manipulation behaviors.
The 2HER extension addresses this limitation by creating a second type of virtual goal in addition to the conventional HER goal based on the object’s future positions. Specifically, while standard HER generates goals from states the object actually reaches, the 2HER mechanism also generates goals from the future positions of the agent’s end-effector, substituting them for the corresponding object position. In summary, the agent is retrospectively rewarded for having moved the manipulator to a location where the object could have been.

By combining the rewards from two distinct hindsight goals, one for agent-object interaction and another for task completion, we create a composite learning signal. This signal incentivizes the agent to engage with the object, accelerating the discovery of interaction dynamics while simultaneously learning to transport the object to its final destination.
Thus, the reward function, which applies to both real transitions and hindsight relabeling, is composed of two sparse terms, as formalized in Equation~\ref{equation:reward_2her}.
The task completion reward, \( r_{\text{goal}}(s_{t+1}) \), is defined as \( 0 \) if \( d(p_{\mathrm{obj}}(s_{t+1}), g) \le \epsilon \), and \( -1 \) otherwise, where \( g \) denotes the goal position and \( p \) denotes position. Similarly, the interaction reward, \( r_{\text{object}}(s_{t+1}) \), is defined as \( 0 \) if \( d(p_{\mathrm{agent}}(s_{t+1}), p_{\mathrm{obj}}(s_{t+1})) \le \epsilon \), and \( -1 \) otherwise.
In this context, $\epsilon$ refers to the object's proximity to the target, and is intrinsic to the environment. This value can correspond to a physical distance, such as meters for environments like Fetch robotics, or even pixels for tasks like object tracking.
The hyperparameter $Cr$, which takes values in the range $[0, 1]$, balances the relative importance of these two terms in the final reward.

\begin{equation}
    r(s_t, a_t, s_{t+1}) = (1 - C_r) \cdot r_{\text{goal}}(s_{t+1}) + C_r \cdot r_{\text{object}}(s_{t+1}) 
    \label{equation:reward_2her}
\end{equation}





Additionally, to enhance the learning efficiency and system stability, we introduce two additional modifications to the original HER algorithm. First, we apply a selective trajectory filter, restricting hindsight relabeling to episodes that exhibit meaningful interaction, identified by changes in the \textit{achieved goal} between the initial and final position. This focuses learning on transitions that contain meaningful interactions. Formally, a trajectory \( \tau = (s_0, s_1, ..., s_T) \) is considered valid for hindsight relabeling if it satisfies the condition $d(p_{\mathrm{obj}}(s_T) - p_{\mathrm{obj}}(s_0)) > \delta$, where \( \delta \) is the minimum threshold for the object's displacement.

Second, we modified the behavior of the hyperparameter $k$, which determines the number of new goals sampled for each transition, allowing its value to vary over the course of training.
The value of $k$ can be selected from the range $[0, T - t]$, where $k=0$ disables HER and the upper bound represents the number of remaining steps in the episode.
High values of $k$ accelerate learning but tend to cause instability during the training process due to the low proportion of real data, as experimentally demonstrated by ~\citet{NIPS2017_453fadbd}. Therefore, the value of $k$ is initialized to a high value and gradually decays over the course of training, allowing robust learning without sacrificing the initial benefit of a high relabeling rate.


Finally, the revisited HER buffer is integrated into the buffer containing the experiences of the original MOC iterations. Once this combined buffer reaches the size defined by the batch size hyperparameter, the data is used to improve each option. For this, a corresponding batch of experience is first shuffled and then partitioned into non-overlapping mini-batches. The option's policies are subsequently updated by performing a gradient step on each mini-batch. After all options have been updated, both the HER and MOC buffers are cleared to be repopulated as the training continues. Our full code is available at \href{https://github.com/ramos-ai/MOC_2HER}{https://github.com/ramos-ai/MOC\_2HER}.

\section{Experiments}

We now present empirical results to demonstrate that our approach:
(i) handles multi-goal environments with sparse rewards that MOC is unable to solve, 
(ii) effectively solves object manipulation tasks by employing 2HER, 
and (iii) preserves the diversity and quality of options.

\subsection{Methodology}


This work aims to enhance option discovery in environments with continuous action spaces and sparse rewards. To this end, we evaluate MOC-HER and MOC-2HER algorithms in the Fetch tasks from Gymnasium Robotics~\citep{plappert2018multigoalreinforcementlearningchallenging}. Their performance, configured with 2, 4, and 8 options, is benchmarked against standard MOC~\citep{NEURIPS2021_24cceab7} and IOC~\citep{Khetarpal_Klissarov_Chevalier-Boisvert_Bacon_Precup_2020} baselines, which were run within the same experimental setup for a direct comparison. Furthermore, to demonstrate the scalability of our contribution, we integrated HER and 2HER into the IOC algorithm, creating a new baseline for comparison against our approaches. The source code of all algorithms and experiments is publicly available at \href{https://github.com/ramos-ai/MOC_2HER}{https://github.com/ramos-ai/MOC\_2HER}.

We begin our evaluation in the FetchReach environment~\citep{plappert2018multigoalreinforcementlearningchallenging}, a multi-goal scenario that requires an industrial robot to reach a pre-defined target location.
The environment features a continuous action space with four possible actions, each ranging from $-1$ to $1$. The observation space contains 16 features, with three values representing the achieved goal and three more representing the desired goal, all of which can take values in the range $-\infty$ to $\infty$.
In the Gymnasium Robotics environments~\mbox{\citep{gymnasium_robotics2023github}}, the reward is $0$ when the Euclidean distance between the robot’s achieved and desired goal positions is below $0.05$; otherwise, it is $-1$. This sparse reward formulation is expressed in Equation~\ref{equation:reward_sparse_withvalues}.

\begin{eqnarray}
    r_{t} =
    \begin{cases}
    0, & \text{if } \hspace{5pt} \| \text{achieved\_goal} - \text{desired\_goal} \|_2 < 0.05 \\
    -1, & \text{otherwise}
    \end{cases}
    \label{equation:reward_sparse_withvalues}
\end{eqnarray}

The remaining environments, FetchPush, FetchSlide, and FetchPickAndPlace~\citep{plappert2018multigoalreinforcementlearningchallenging}, introduce object manipulation tasks that significantly increase complexity. In FetchPush, the agent must move a box on the table surface to a target location. FetchSlide extends this challenge by placing the target beyond the robot's direct reach, requiring the agent to apply sufficient force for the box to slide to the goal. FetchPickAndPlace is the most complex scenario, demanding that the agent not only reaches for the box but also grasps, lifts, and transports it to a target position that may be elevated above the surface. 
For all object-centric tasks, the reward remains the same as in the FetchReach scenario (Equation~\ref{equation:reward_sparse_withvalues}), except that the achieved goal feature refers to the position of the box instead of the agent's position.
Consequently, the observation space is expanded to include the state of the object, and for FetchPickAndPlace, the action space is augmented to incorporate gripper control. As a result, the overall observation space consists of 31 features, including both the achieved and desired goal positions. In the experimental setup, the object's initial position is randomized once at the beginning and then remains fixed at the beginning of all subsequent episodes, while the goal is resampled for each new episode.

Each episode consists of 50 time steps and never terminates before this predefined limit, requiring the algorithm to not only guide the robot to the target position but also to keep it there. 
The training process is divided into cycles, or iterations, where each iteration consists of collecting data for 2000 time steps, or 40 complete episodes, followed by a model training and optimization step.

Since the initial object-to-goal distance varies between episodes, the cumulative reward may not be a clear indicator of learning progress. Consequently, we evaluate performance based on the percentage of successful episodes. As defined in ~\citep{NIPS2017_453fadbd}, an episode is deemed successful if the distance between the object and the goal at the end of the episode is below 7 cm for FetchPush and FetchPickAndPlace tasks, and below 20 cm for the FetchSlide environment.

A key feature of the MOC framework is its simultaneous update mechanism, which provides a learning signal to multiple options concurrently. This design accelerates skill discovery and ensures that the full set of options remains active and useful throughout training. Thus, we investigate whether this property is maintained after the integration with HER. We also extend this analysis to the IOC framework to directly compare how both architectures handle option utilization under our proposed enhancements. To quantify this behavior, we report the utilization rate of the options during agent-environment interactions.

\subsection{Numerical results}

\begin{figure}[h]
    \centering
    \setlength{\abovecaptionskip}{0pt}
    \setlength{\belowcaptionskip}{0pt}
    \includegraphics[width=0.95\linewidth]{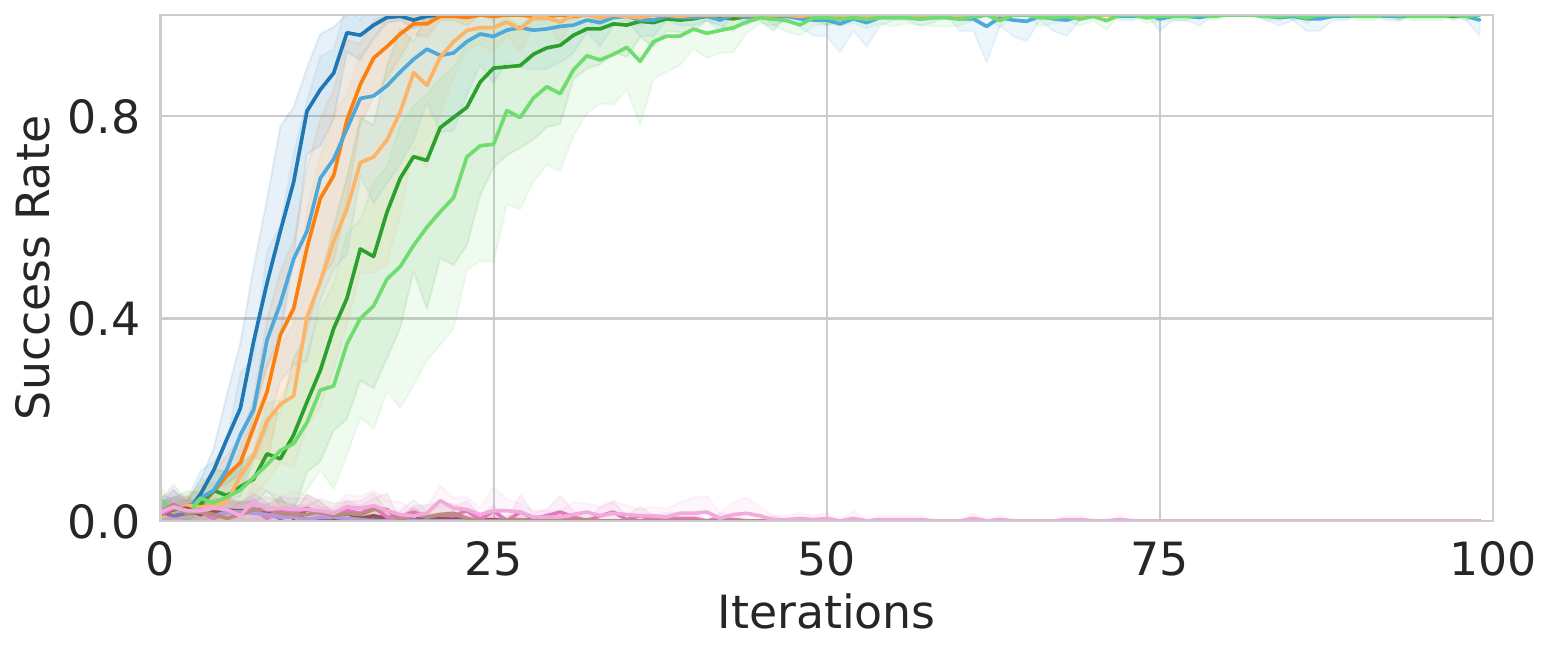}        
    \vspace{0.25em}
    \scalebox{1}{
        \begin{minipage}{1\linewidth}
            \centering
            \footnotesize
            \renewcommand{\arraystretch}{1.2}
            \begin{tabular}{@{}p{0.33\linewidth}@{\hspace{0.5em}} p{0.33\linewidth}@{\hspace{0.5em}} p{0.33\linewidth}@{}}
                \textcolor{reach_moc2her2}{\rule[0.5ex]{2em}{1.5pt}} MOC-HER 2opt &
                \textcolor{reach_moc2her4}{\rule[0.5ex]{2em}{1.5pt}} MOC-HER 4opt &
                \textcolor{reach_moc2her8}{\rule[0.5ex]{2em}{1.5pt}} MOC-HER 8opt \\
                \textcolor{reach_ioc2her2}{\rule[0.5ex]{2em}{1.5pt}} IOC-HER 2opt &
                \textcolor{reach_ioc2her4}{\rule[0.5ex]{2em}{1.5pt}} IOC-HER 4opt &
                \textcolor{reach_ioc2her8}{\rule[0.5ex]{2em}{1.5pt}} IOC-HER 8opt \\
                \textcolor{reach_moc2}{\rule[0.5ex]{2em}{1.5pt}} MOC 2opt &
                \textcolor{reach_moc4}{\rule[0.5ex]{2em}{1.5pt}} MOC 4opt &
                \textcolor{reach_moc8}{\rule[0.5ex]{2em}{1.5pt}} MOC 8opt \\
                \textcolor{reach_ioc2}{\rule[0.5ex]{2em}{1.5pt}} IOC 2opt &
                \textcolor{reach_ioc4}{\rule[0.5ex]{2em}{1.5pt}} IOC 4opt &
                \textcolor{reach_ioc8}{\rule[0.5ex]{2em}{1.5pt}} IOC 8opt
            \end{tabular}
        \end{minipage}
    }
    \vspace{0.3em}        
    \caption{Comparison of different approaches with 2, 4 and 8 options in FetchReach over 100 iterations. Solid line and shaded region show the mean and standard deviation of success rates across 10 seeds.}    
    \label{fig:FetchReach}
\end{figure}

Our first experiment consists of comparing the results of MOC-HER and IOC-HER with those obtained with the standard algorithms in the FetchReach environment. Since this task does not involve object manipulation, the 2HER component is not activated.
The results are presented in Figure~\ref{fig:FetchReach}, where the shaded areas represent the standard deviation across multiple runs. Results with HER extensions achieved success rates between 99\% and 100\% in the final iteration.
Furthermore, the insights demonstrate that the MOC and IOC algorithms, regardless of the number of options, failed to identify a policy capable of solving the environment during the analyzed training period. Their results remained close to 0 at the end of training. 
Both MOC-HER and IOC-HER solved the task with substantially improved efficiency, although MOC-HER reached a 100\% success rate, on average, 32.6\% more rapidly.
It is noted that the number of time steps required to solve the environment increases according to the number of options.
The obtained results highlight the benefits of implementing the HER buffer and the potential advantages of trajectory reward recalculation.

\begin{figure}[h]
    \centering
    \setlength{\abovecaptionskip}{0pt}
    \setlength{\belowcaptionskip}{0pt}
    \includegraphics[width=0.95\linewidth]{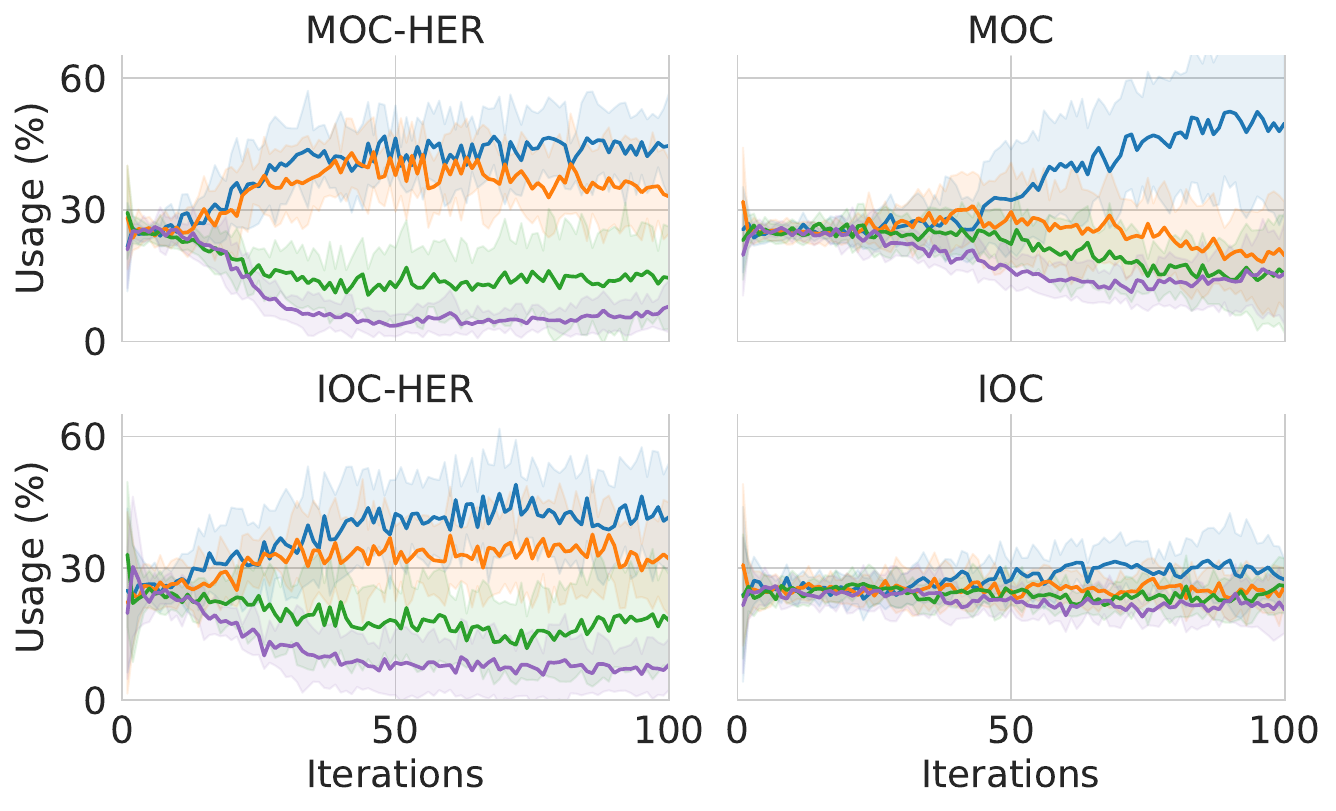}        
    \vspace{-0.3em}    
    \begin{minipage}{\linewidth}
        \centering
        \footnotesize
        \scalebox{1}{
            \textcolor{tab_option1}{\rule[0.5ex]{2em}{1.5pt}} Option1 \qquad
            \textcolor{tab_option2}{\rule[0.5ex]{2em}{1.5pt}} Option2 \qquad
            \textcolor{tab_option3}{\rule[0.5ex]{2em}{1.5pt}} Option3 \qquad
            \textcolor{tab_option4}{\rule[0.5ex]{2em}{1.5pt}} Option4
        }
    \end{minipage}    
    \vspace{0.3em}
    \caption{Utilization rates for each option in the 4-option configuration for the FetchReach task, over 100 iterations. Solid line and shaded region show the mean and standard deviation across 10 seeds.}
    \label{fig:FetchReachOptions}
\end{figure}

 Figure~\ref{fig:FetchReachOptions} illustrates the usage percentage of each option in the 4 options scenario. For both MOC-HER and IOC-HER, a clear specialization emerges, with some options being selected significantly more frequently than others. The difference between the most and least utilized options ranged from 39.5\% to 9.5\% in MOC-HER, and from 37.8\% to 11.5\% in IOC-HER.
 This imbalance suggests the number of options might be overestimated for this task, though it is noteworthy that all options remain active and contribute to the final policy. 
 Both standard MOC and IOC failed to solve the task, reflected in their unpredictable option usage. MOC's utilization leaned heavily towards one option (35.5\% compared to 18.1\% for the least used), while IOC's usage remained largely static and unspecialized.
 Additional information for the 2-option and 8-option cases can be found in Appendix~\ref{sec:additional_quantitative_results}.

\begin{figure*}[t]
    \centering
    \setlength{\abovecaptionskip}{0pt} 
    \setlength{\belowcaptionskip}{0pt} 
    \includegraphics[width=0.95\linewidth]{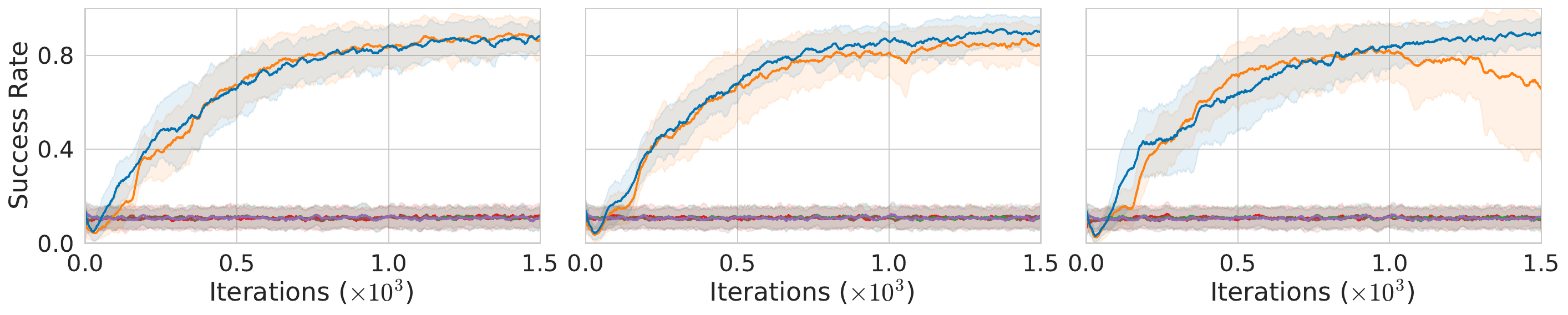}    
    \vspace{0em}
    \begin{minipage}{0.88\linewidth}
        \centering
        \begin{tabular}{p{0.33\linewidth} p{0.33\linewidth} p{0.33\linewidth}}
            \centering \small(a) FetchPush 2 options &
            \centering \small(b) FetchPush 4 options &
            \centering \small(c) FetchPush 8 options
        \end{tabular}
    \end{minipage}   
    
    \vspace{0.25em}
    \begin{minipage}{0.8\linewidth}
        \centering
        \small
        \scalebox{0.8}{
            \textcolor{ioc}{\rule[0.5ex]{2em}{1.5pt}} IOC \quad
            \textcolor{iocher}{\rule[0.5ex]{2em}{1.5pt}} IOC-HER \quad
            \textcolor{ioc2her}{\rule[0.5ex]{2em}{1.5pt}} IOC-2HER \quad
            \textcolor{moc}{\rule[0.5ex]{2em}{1.5pt}} MOC \quad
            \textcolor{mocher}{\rule[0.5ex]{2em}{1.5pt}} MOC-HER \quad
            \textcolor{moc2her}{\rule[0.5ex]{2em}{1.5pt}} MOC-2HER
        }
    \end{minipage}
    
    \vspace{0.5em}
    \caption{Comparison of different approaches with 2, 4 and 8 options in FetchPush over $1.5\times10^{3}$ iterations. Solid line and shaded region show the mean and standard deviation of success rates across 5 seeds. Results are averaged over a window size of 20 iterations.}
    \label{fig:FetchPush_learning}
\end{figure*}

\begin{figure*}[t]
    \centering
    \setlength{\abovecaptionskip}{0pt} 
    \setlength{\belowcaptionskip}{0pt} 
    \includegraphics[width=0.95\linewidth]{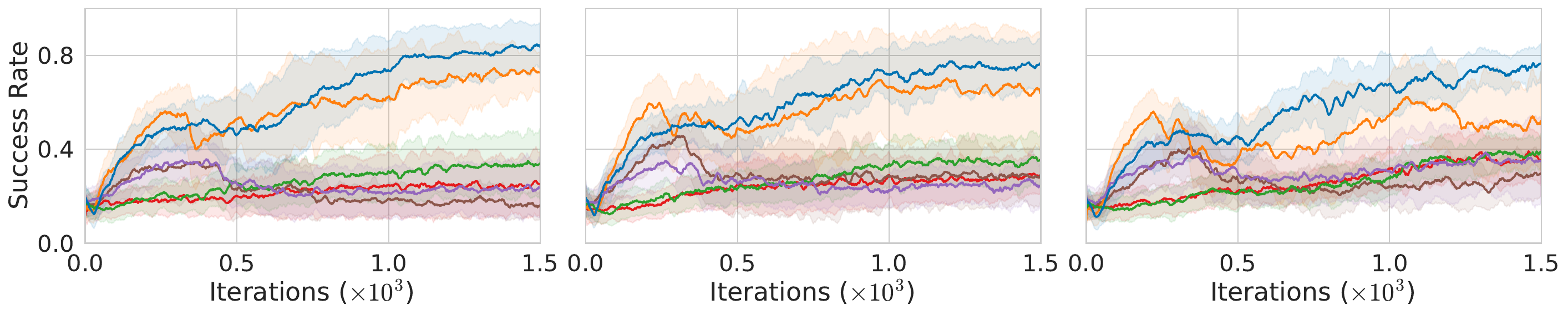}    
    \vspace{0em}
    \begin{minipage}{0.88\linewidth}
        \centering
        \begin{tabular}{p{0.33\linewidth} p{0.33\linewidth} p{0.33\linewidth}}
            \centering \small(a) FetchSlide 2 options &
            \centering \small(b) FetchSlide 4 options &
            \centering \small(c) FetchSlide 8 options
        \end{tabular}
    \end{minipage}   
    
    \vspace{0.25em}
    \begin{minipage}{0.8\linewidth}
        \centering
        \small
        \scalebox{0.8}{
            \textcolor{ioc}{\rule[0.5ex]{2em}{1.5pt}} IOC \quad
            \textcolor{iocher}{\rule[0.5ex]{2em}{1.5pt}} IOC-HER \quad
            \textcolor{ioc2her}{\rule[0.5ex]{2em}{1.5pt}} IOC-2HER \quad
            \textcolor{moc}{\rule[0.5ex]{2em}{1.5pt}} MOC \quad
            \textcolor{mocher}{\rule[0.5ex]{2em}{1.5pt}} MOC-HER \quad
            \textcolor{moc2her}{\rule[0.5ex]{2em}{1.5pt}} MOC-2HER
        }
    \end{minipage}    
    
    \vspace{0.5em}
    \caption{Comparison of different approaches with 2, 4 and 8 options in FetchSlide over $1.5\times10^{3}$ iterations. Solid line and shaded region show the mean and standard deviation of success rates across 5 seeds. Results are averaged over a window size of 20 iterations.}
    \label{fig:FetchSlide_learning}
\end{figure*}

\begin{figure*}[t]
    \centering
    \setlength{\abovecaptionskip}{0pt} 
    \setlength{\belowcaptionskip}{0pt} 
    \includegraphics[width=0.95\linewidth]{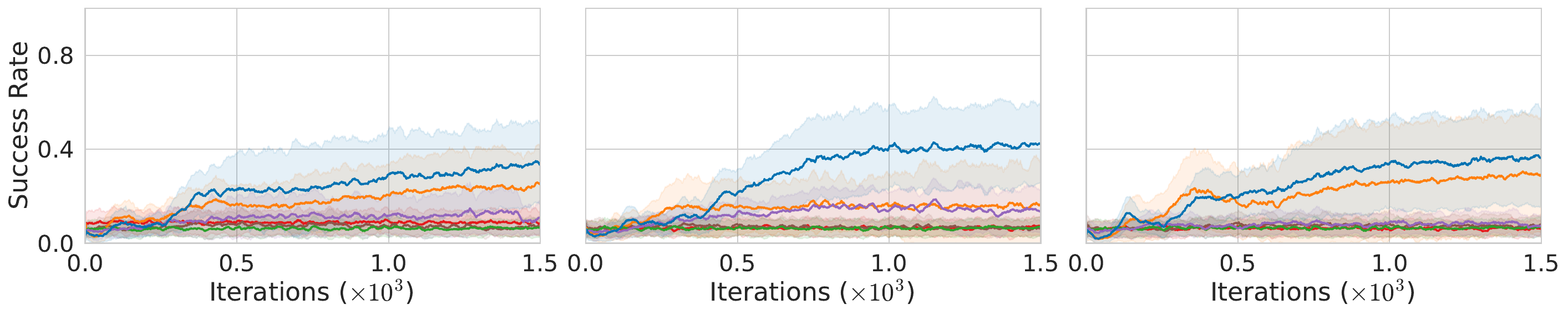}    
    \vspace{0em}
    \begin{minipage}{0.88\linewidth}
        \centering
        \begin{tabular}{p{0.33\linewidth} p{0.33\linewidth} p{0.33\linewidth}}
            \centering \small(a) FetchPickAndPlace 2 options &
            \centering \small(b) FetchPickAndPlace 4 options &
            \centering \small(c) FetchPickAndPlace 8 options
        \end{tabular}
    \end{minipage}   
    
    \vspace{0.25em}
    \begin{minipage}{0.8\linewidth}
        \centering
        \small
        \scalebox{0.8}{
            \textcolor{ioc}{\rule[0.5ex]{2em}{1.5pt}} IOC \quad
            \textcolor{iocher}{\rule[0.5ex]{2em}{1.5pt}} IOC-HER \quad
            \textcolor{ioc2her}{\rule[0.5ex]{2em}{1.5pt}} IOC-2HER \quad
            \textcolor{moc}{\rule[0.5ex]{2em}{1.5pt}} MOC \quad
            \textcolor{mocher}{\rule[0.5ex]{2em}{1.5pt}} MOC-HER \quad
            \textcolor{moc2her}{\rule[0.5ex]{2em}{1.5pt}} MOC-2HER
        }
    \end{minipage}    
    
    \vspace{0.5em}
    \caption{Comparison of different approaches with 2, 4 and 8 options in FetchPickAndPlace over $1.5\times10^{3}$ iterations. Solid line and shaded region show the mean and standard deviation of success rates across 5 seeds. Results are averaged over a window size of 20 iterations.}
    \label{fig:FetchPickAndPlace_learning}
\end{figure*}

Our subsequent experiments involve object manipulation tasks, where our 2HER extension is activated. In this stage, we compare the performance of MOC-2HER and IOC-2HER against the standard MOC and IOC baselines, as well as their extensions with the simple HER (MOC-HER and IOC-HER). The results for the FetchPush, FetchSlide, and FetchPickAndPlace tasks are presented in Figures~\ref{fig:FetchPush_learning},~\ref{fig:FetchSlide_learning}, and~\ref{fig:FetchPickAndPlace_learning}, respectively. Each figure displays the learning curves for configurations with 2, 4, and 8 options. In Appendix~\ref{sec:appendix_implementation_details}, we provide a complete list of hyperparameters and configurations used in the experiments.

IOC, IOC-HER, MOC, and MOC-HER demonstrated limited performance in all the tested tasks, with final success rates remaining below 12\% and 13\% in FetchPush and FetchPickAndPlace, respectively. The best performance of these baseline algorithms was in the FetchSlide scenario, achieving a success rate of up to 39\%. However, this result was significantly lower than the performance of the 2HER-augmented versions.
In contrast, the algorithms augmented with 2HER consistently outperformed both their standard versions and their HER counterparts in all tested scenarios. In the FetchPush environment, MOC-2HER and IOC-2HER achieved high success rates, concluding training with final score of up to 90\% and 87\%, respectively, across the different option configurations (2, 4, and 8). On FetchSlide, MOC-2HER achieved success rates between 76\% and 84\%, while IOC-2HER ranged between 66\% and 87\%. In the more challenging FetchPickAndPlace task, only MOC-2HER achieved a meaningful success rate, reaching up to 42\%, whereas even IOC-2HER reached less than 29\%. 
These results empirically validate the efficacy of the 2HER mechanism. This holds not only for MOC, which was the primary focus of this work, but also demonstrates its generalizability when integrated with other HRL off-policy algorithms such as IOC.

A notable effect of 2HER strategy can be observed during the initial training phase. In the initial steps, the algorithm gives greater preference to interacting with objects, a behavior that frequently pushes the object away from the goal. This leads to a temporary dip in performance, which is quickly corrected as the agent learns to correlate object interaction with successful goal achievement. Furthermore, as the number of options is increased, IOC-2HER tends to exhibit greater instability than MOC-2HER. This is particularly evident in the 8-option FetchPush and FetchSlide scenarios, where the success rate for IOC-2HER began to decline during the final stages of training.

\begin{figure*}[t]
    \centering
    \setlength{\abovecaptionskip}{0pt} 
    \setlength{\belowcaptionskip}{0pt} 
    
    \begin{minipage}[b]{0.329\textwidth}
        \centering
        \includegraphics[width=\linewidth]{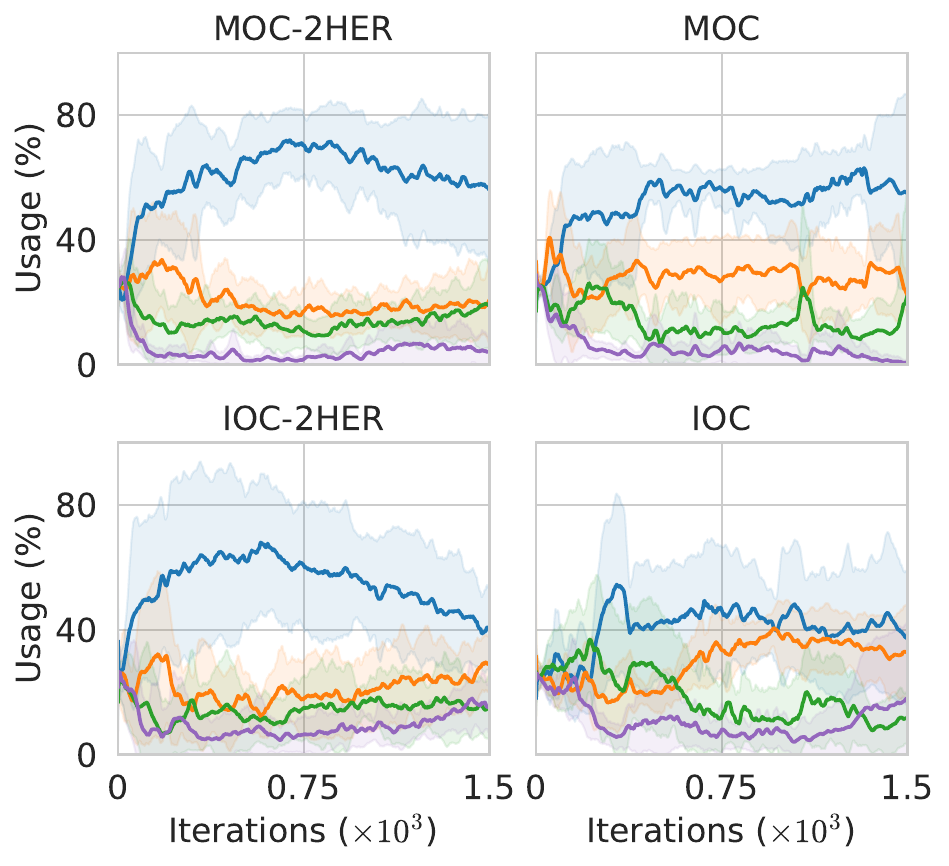}
        \par\centering {\small(a) FetchPush}
    \end{minipage}
    \hfill
    \begin{minipage}[b]{0.329\textwidth}
        \centering
        \includegraphics[width=\linewidth]{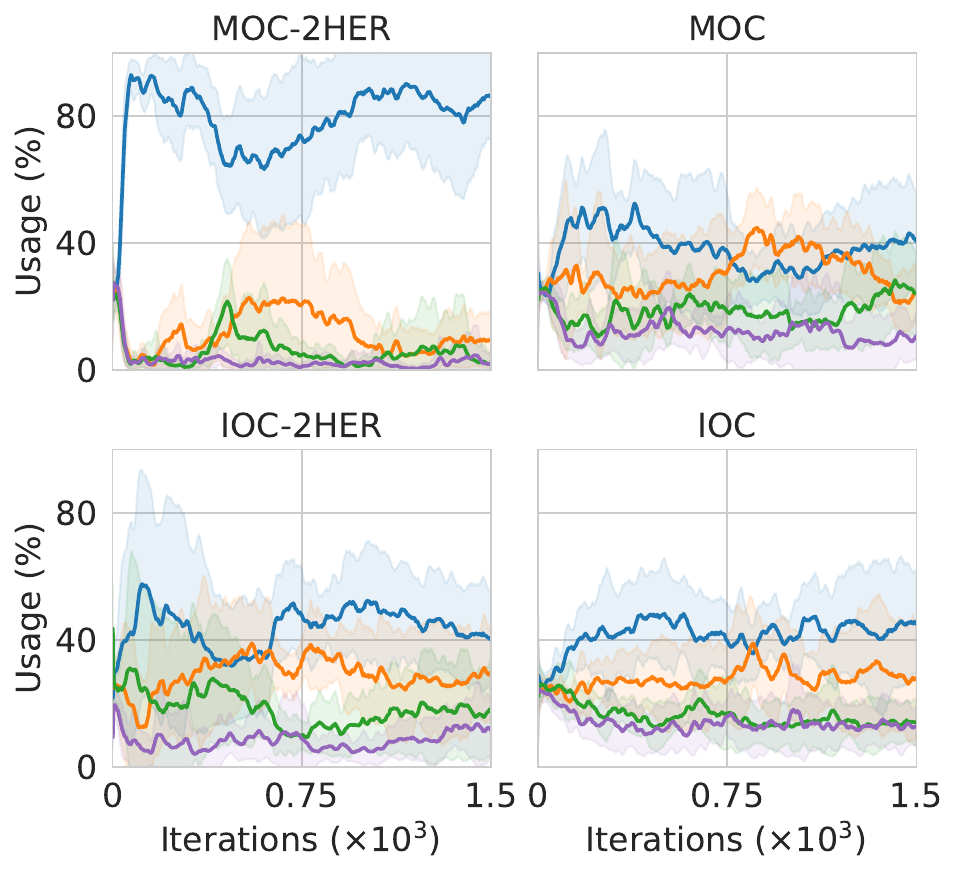}
        \par\centering {\small(b) FetchSlide}
    \end{minipage}
    \hfill
    \begin{minipage}[b]{0.329\textwidth}
        \centering
        \includegraphics[width=\linewidth]{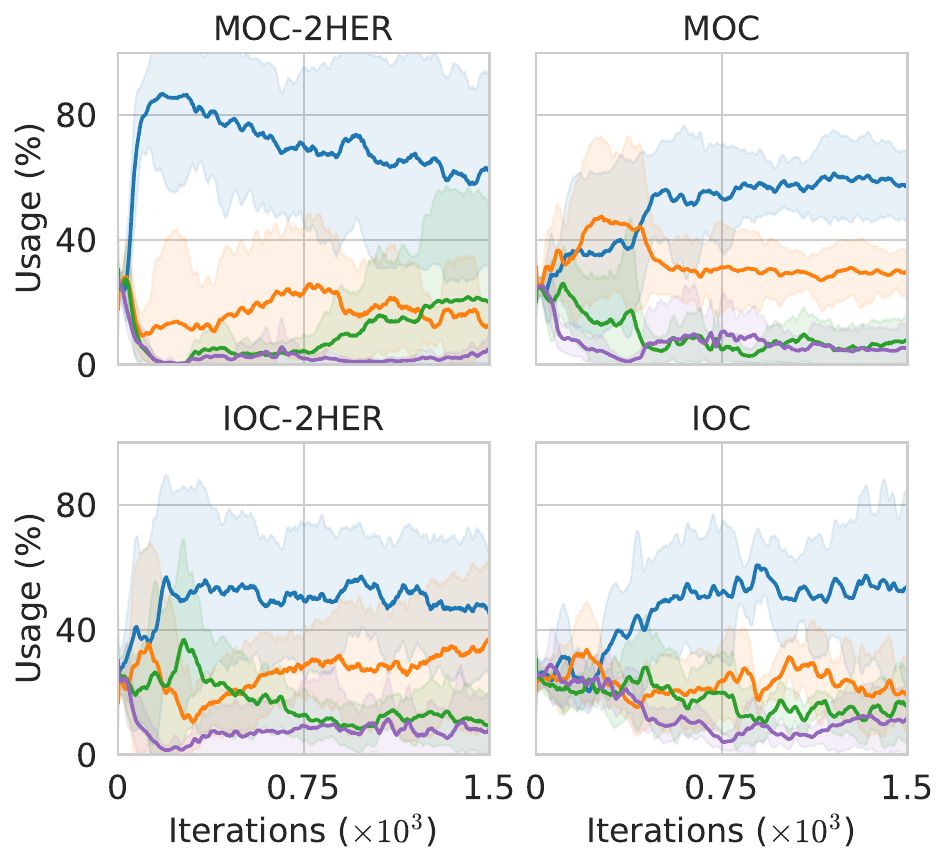}
        \par\centering {\small(c) FetchPickAndPlace}
    \end{minipage}

    \begin{minipage}{0.8\linewidth}
        \centering
        \scalebox{0.75}{
            \begin{tikzpicture}
                \draw[line width=1.5pt, color=tab_option1]   (0,0) -- (0.5,0) node[right,black] {Option1};
                \draw[line width=1.5pt, color=tab_option2] (3,0) -- (3.5,0) node[right,black] {Option2};
                \draw[line width=1.5pt, color=tab_option3] (6,0) -- (6.5,0) node[right,black] {Option3};
                \draw[line width=1.5pt, color=tab_option4] (9,0) -- (9.5,0) node[right,black] {Option4};
            \end{tikzpicture}
        }
    \end{minipage}
    
    \caption{Option utilization rates for the 4-option configuration across various tasks and approaches over $1.5\times10^{3}$ iterations. Solid line and shaded region show the mean and standard deviation across 5 seeds. Results are averaged over a window size of 20 iterations.}
    \label{fig:OptionUsage_2HER}
\end{figure*}

Figure~\ref{fig:OptionUsage_2HER} presents the option utilization rates during training on the object interaction tasks for the 4 options setting. We observe that 2HER augmented agents, at the end of process, consistently prioritized a small subset of options, with the most used option selected between 53.6-83.1\% of the time in MOC-2HER and 41.3-43.9\% in IOC-2HER. 
In contrast, without 2HER extension, option usage remained highly unstable throughout training, reflecting the agent's inability to learn a meaningful policy from the sparse reward signal. Aligned with the learning curve results, this indicates that multi-option learning, particularly in sparse reward scenarios, is more effective for teaching the agent a policy and for specializing options for specific tasks than maintaining balanced utilization. This is supported by the superior performance of MOC-2HER over IOC-2HER, even in scenarios where IOC-2HER exhibited a more balanced option utilization. Despite this, all options were employed at some stage of the learning process, with the least-favored option reaching 2.4–7.8\% in MOC-2HER and 9.5–12.1\% in IOC-2HER. Option usage rates for settings with different numbers of options are provided in Appendix~\ref{sec:additional_quantitative_results}.

Table~\ref{tab:all_algorithms_results} presents the final success rates for all evaluated algorithms across the four Fetch robotics tasks, complementing the learning curves shown in Figures~\ref{fig:FetchReach},~\ref{fig:FetchPush_learning},~\ref{fig:FetchSlide_learning},~\ref{fig:FetchPickAndPlace_learning}.
We observe that the standard MOC and IOC struggled to solve the tasks, achieving average success rates between 11\% and 14\%. In contrast, algorithms augmented with HER and 2HER (IOC-HER, IOC-2HER, MOC-HER, and MOC-2HER) achieved the best performance with a 2-option configuration, where average success rates approached 100\% in some tasks. These results effectively demonstrate the benefits of HER for positioning tasks and 2HER for object manipulation.
When the number of options is increased, performance gradually declines, with higher variability observed for 8 options. Overall, our results emphasize that HRL algorithms augmented with HER and 2HER and a moderate number of options (2-4) achieve the best balance between performance and stability. Furthermore, by comparing the mean success rate across environments, we highlight the advantage of MOC-2HER with 4 options (69\%) over the other methods.

\begin{table}[t]
    \centering
    \setlength{\abovecaptionskip}{2.5pt}
    \setlength{\belowcaptionskip}{0pt}
    \caption{Final success rates for all evaluated algorithms across the four Fetch robotics tasks. The table reports the mean, standard deviation, and overall mean for each approach in the 2, 4, and 8-option configurations from our experiments. A dash indicates scenarios where an algorithm is not applicable.}
    \resizebox{1\linewidth}{!}{%
    \begin{tabular}{l c c c c c c}
        \hline
        \rule{0pt}{2ex}
        \textbf{Algorithm} & \textbf{Options} &
        \textbf{Reach} & \textbf{Push} & \textbf{Slide} & \textbf{PickAndPlace} & \textbf{Mean} \\
        \hline
        \rule{0pt}{2ex}                 
        IOC      & 2 & 0.00 (0) & 0.12 (0.05) & 0.25 (0.14) & 0.09 (0.05) & 0.11 \\
                 & 4 & 0.00 (0) & 0.11 (0.04) & 0.28 (0.07) & 0.07 (0.04) & 0.11 \\
                 & 8 & 0.00 (0) & 0.10 (0.04) & 0.35 (0.10) & 0.06 (0.04) & 0.13 \\[2pt]
                 
        IOC-HER & 2 & 0.99 (0.03) & 0.11 (0.04) & 0.16 (0.06) & 0.07 (0.04) & 0.33 \\
                 & 4 & \textbf{1.00 (0)} & 0.11 (0.06) & 0.29 (0.12) & 0.06 (0.03) & 0.36 \\
                 & 8 & \textbf{1.00 (0)} & 0.11 (0.05) & 0.39 (0.10) & 0.08 (0.04) & 0.39 \\[2pt]
                 
        IOC-2HER & 2 & - & 0.87 (0.08) & 0.73 (0.08) & 0.25 (0.16) & 0.62 \\
                 & 4 & - & 0.84 (0.09) & 0.64 (0.24) & 0.16 (0.18) & 0.55 \\
                 & 8 & - & 0.66 (0.31) & 0.52 (0.21) & 0.29 (0.25) & 0.49 \\[2pt]
                 
        MOC      & 2 & 0.00 (0) & 0.11 (0.05) & 0.34 (0.15) & 0.07 (0.04) & 0.13 \\
                 & 4 & 0.00 (0) & 0.11 (0.05) & 0.35 (0.11) & 0.07 (0.04) & 0.13 \\
                 & 8 & 0.00 (0) & 0.11 (0.05) & 0.39 (0.08) & 0.07 (0.04) & 0.14 \\[2pt]
                 
        MOC-HER & 2 & \textbf{1.00 (0)} & 0.11 (0.05) & 0.24 (0.13) & 0.11 (0.07) & 0.36 \\
                 & 4 & \textbf{1.00 (0)} & 0.11 (0.05) & 0.24 (0.10) & 0.13 (0.09) & 0.37 \\
                 & 8 & \textbf{1.00 (0)} & 0.11 (0.05) & 0.35 (0.18) & 0.08 (0.04) & 0.38 \\[2pt]
                 
        MOC-2HER & 2 & - & 0.88 (0.06) & \textbf{0.84 (0.10)} & 0.33 (0.17) & 0.68 \\
                 & 4 & - & 0.90 (0.07) & 0.76 (0.11) & \textbf{0.42 (0.16)} & \textbf{0.69} \\
                 & 8 & - & \textbf{0.90 (0.06)} & 0.77 (0.09) & 0.36 (0.20) & 0.68 \\
        \hline
    \end{tabular}%
    }
    \label{tab:all_algorithms_results}
\end{table}

\section{Conclusion}

In this paper, we addressed the challenge of applying HRL in multi-goal environments with sparse rewards. We first introduced MOC-HER, an integration of HER with the MOC framework, which proved effective in solving tasks that previously failed under standard HRL algorithms. To tackle the more complex problem of object manipulation, we proposed Dual Objectives Hindsight Experience Replay (2HER), a novel mechanism that generates virtual goals for both the object's state and the agent's effector. Our experiments in robotic environments demonstrate that MOC-2HER successfully learns complex manipulation skills, solving tasks where both the original MOC and our MOC-HER baseline fail. Thus, we achieved an average success rate of up to 90\% in the tested scenarios, while both MOC-HER and standard MOC achieved less than 12\%. Furthermore, our analysis indicates that, in sparse reward scenarios, MOC’s multi-option update mechanism contributes primarily to enhanced performance, rather than to balanced option usage. To demonstrate the scalability of our approach, we also integrated 2HER with the IOC algorithm, yielding similar significant performance improvements.

Despite these results, it is important to acknowledge the limitations of our approach. The experimental setup was restricted to a static object placement, where the object's position was randomized once at the start of each training run and then remained fixed for all subsequent episodes.
In addition, our method's reliance on HER makes it inherently compatible only with off-policy algorithms. Finally, the framework is designed exclusively for goal-based environments, limiting its applicability to tasks where success can be clearly defined by reaching a target state.
Thus, the limitations of our current study naturally motivate several directions for future research. While this work has focused on robotics manipulation, extending 2HER to other domains, such as navigation, dynamic-object interaction or multi-agent coordination, represents an important next step. Since our method relies on a fixed number of options, integrating adaptive strategies like Dynamic Option Creation~\citep{Melchiades:2025} could enable the agent to learn a more autonomous and efficient set of options.
Another promising avenue for future work is the transferability of learned options between different tasks.

\begin{acks}
We thank the anonymous reviewers for their valuable feedback. This work was supported by Kunumi Institute. The authors thank the institution for its financial support and commitment to advancing scientific research. This research was also partially supported by Conselho Nacional de Desenvolvimento Científico e Tecnológico - CNPq (grants 313845/2023-9, 443184/2023-2, 445238/2024-0, and 404800/2025-4).
\end{acks}


\balance

\bibliographystyle{ACM-Reference-Format}
\bibliography{sample}

\newpage  

\onecolumn

\appendix

\section{Implementation details} \label{sec:appendix_implementation_details}

We selected a set of hyperparameter configuration for each environment based on preliminary experiments and applied it consistently across all algorithm implementations to enable a fair comparison. Table~\ref{tab:selected_hyperparameters} presents the values used in our experiments, listing only the hyperparameters added by our approach and those from the standard algorithm that showed more relevance to our implementation.

\begin{table}[hb]
    \centering
    \setlength{\abovecaptionskip}{2.5pt} 
    \caption{Hyperparameter configurations for all environments. Key parameters were tuned per environment for each algorithm, with a dash denoting non-applicable parameters. Unlisted parameters retained their default values.}
    \resizebox{0.75\linewidth}{!}{%
        \begin{tabular}{llcccccc} 
            \hline
            \rule{0pt}{2ex}\textbf{Environment} & \textbf{Hyperparameter} & \textbf{MOC-2HER} & \textbf{MOC-HER} & \textbf{MOC} & \textbf{IOC-2HER} & \textbf{IOC-HER} & \textbf{IOC} \\
            \hline
            \rule{0pt}{2.5ex}\textbf{Reach} & k \rule{0pt}{2ex}       & -     & 4     & -     & -     & 4     & -     \\
                           & k decay                & -     & -     & -     & -     & -     & -     \\
                           & Cr     & -     & -     & -     & -     & -     & -     \\
                           & 2HER disable               & -     & -     & -     & -     & -     & -     \\
                           & entropy coefficient    & -     & 0     & 0     & -     & 0     & 0     \\
                           & learning rate          & -     & $1\times10^{-4}$     & $1\times10^{-4}$     & -     & $1\times10^{-4}$     & $1\times10^{-4}$     \\
            \hline
            \rule{0pt}{2.5ex}\textbf{Push} & k \rule{0pt}{2ex}       & 8     & 4     & -     & 8     & 4     & -     \\
                           & k decay                & 37     & -     & -     & 37     & -     & -     \\
                           & Cr     & 0.8     & -     & -     & 0.8     & -     & -     \\
                           & 2HER disable               & 150     & -     & -     & 150     & -     & -     \\
                           & entropy coefficient    & 0.005     & 0.005     & 0.005     & 0.005     & 0.005     & 0.005     \\
                           & learning rate          & $1\times10^{-4}$     & $1\times10^{-4}$     & $1\times10^{-4}$     & $1\times10^{-4}$     & $1\times10^{-4}$     & $1\times10^{-4}$     \\
            \hline
            \rule{0pt}{2.5ex}\textbf{Slide} & k \rule{0pt}{2ex}       & 12     & 4     & -     & 12     & 4     & -     \\
                           & k decay                & 33     & -     & -     & 33     & -     & -     \\
                           & Cr     & 0.6     & -     & -     & 0.6     & -     & -     \\
                           & 2HER disable               & 50     & -     & -     & 50     & -     & -     \\
                           & entropy coefficient    & 0.005     & 0.005     & 0.005     & 0.005     & 0.005     & 0.005     \\
                           & learning rate          & $1\times10^{-4}$     & $1\times10^{-4}$     & $1\times10^{-4}$     & $1\times10^{-4}$     & $1\times10^{-4}$     & $1\times10^{-4}$     \\
            \hline
            \rule{0pt}{2.5ex}\textbf{PickAndPlace} & k \rule{0pt}{2ex}       & 4     & 4     & -     & 4     & 4     & -     \\
                           & k decay                & 150     & -     & -     & 150     & -     & -     \\
                           & Cr     & 1     & -     & -     & 1     & -     & -     \\
                           & 2HER disable               & 250     & -     & -     & 250     & -     & -     \\
                           & entropy coefficient    & 0.005     & 0.005     & 0.005     & 0.005     & 0.005     & 0.005     \\
                           & learning rate          & $1\times10^{-4}$     & $1\times10^{-4}$     & $1\times10^{-4}$     & $1\times10^{-4}$     & $1\times10^{-4}$     & $1\times10^{-4}$     \\
        \hline
        \end{tabular}%
    }
    \label{tab:selected_hyperparameters}
\end{table}

The MOC algorithm's training is structured around iterations rather than environment time steps, and its hyperparameter schedules are defined accordingly. Each iteration comprises 2000 time steps, after which a learning step is executed. With an episode length of 50 time steps, each iteration thus encompasses 40 episodes. We detail the key hyperparameters presented in Table~\ref{tab:selected_hyperparameters}: \textit{k} denotes the number of virtual goals sampled per transition for HER/2HER, while \textit{k decay} is the interval (in iterations) at which \textit{k} is decremented by one. The \textit{Cr} coefficient weighs the goal reward ($r_{\text{goal}}$) against the object reward ($r_{\text{object}}$) in the final reward function (Equation~\ref{equation:reward_2her}). The \textit{2HER disable} parameter sets the iteration at which the 2HER module is deactivated. The learning rate and entropy coefficient follow the standard MOC and IOC implementations.

In the PickAndPlace environment, \textit{Cr} is set to $1$, so the 2HER relabeling focuses exclusively on the object interaction reward $r_{\text{object}}$, reflecting that the picking sub-task is the main learning bottleneck. After the \textit{2HER disable} iteration, the 2HER module is deactivated, and the algorithm shifts its focus to the final goal reward $r_{\text{goal}}$ under the standard HER formulation.

\section{Additional quantitative results} \label{sec:additional_quantitative_results}

This appendix provides additional results and analyses to further support the findings presented in the main paper. Specifically, we present empirical values and an analysis of option utilization at the end of training.

\begin{figure}[t]
    \centering
    \setlength{\abovecaptionskip}{0pt} 
    \setlength{\belowcaptionskip}{0pt} 
    \includegraphics[width=\linewidth]{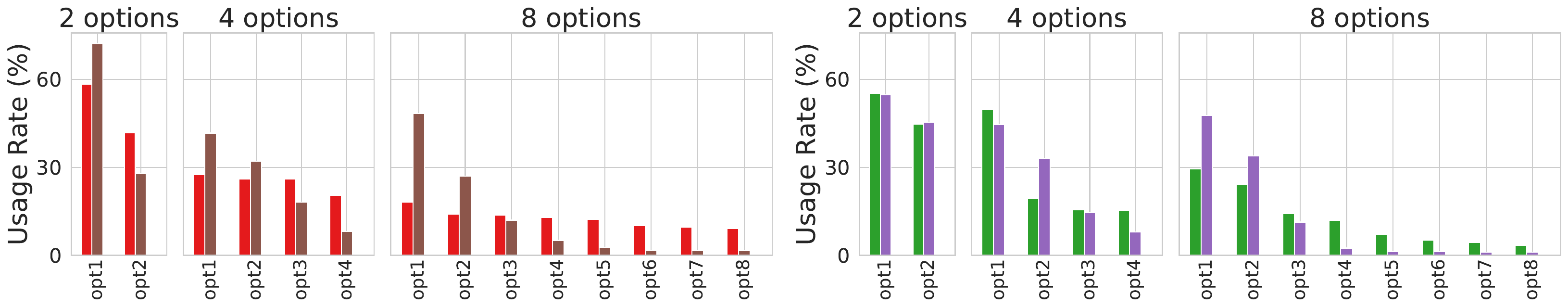}
    
    \begin{minipage}{0.95\linewidth}
        \centering
        \begin{tabular}{p{0.49\linewidth} p{0.49\linewidth}}
            \centering \small(a) FetchReach IOC options &
            \centering \small(b) FetchReach MOC options
        \end{tabular}
    \end{minipage}   

    \vspace{0em}
    \begin{minipage}{0.95\linewidth}
        \centering
        \begin{tabular}{p{0.49\linewidth} p{0.49\linewidth}}
            \centering
            \small
            \scalebox{0.8}{
                \textcolor{ioc}{\rule[0.5ex]{2em}{1.5pt}} IOC \quad
                \textcolor{iocher}{\rule[0.5ex]{2em}{1.5pt}} IOC-HER
            }
            &
            \centering
            \small
            \scalebox{0.8}{
                \textcolor{moc}{\rule[0.5ex]{2em}{1.5pt}} MOC \quad
                \textcolor{mocher}{\rule[0.5ex]{2em}{1.5pt}} MOC-HER
            }
        \end{tabular}
    \end{minipage}
    
    \caption{Option utilization rates for IOC (left panel) and MOC (right panel) methods with 2, 4, and 8 options in the FetchReach task. Bars show the mean usage of each option across 10 seeds after 100 iterations.}
    \label{fig:reach_opts}
\end{figure}

\begin{figure}[t]
    \centering
    \setlength{\abovecaptionskip}{0pt} 
    \setlength{\belowcaptionskip}{0pt} 
    \includegraphics[width=\linewidth]{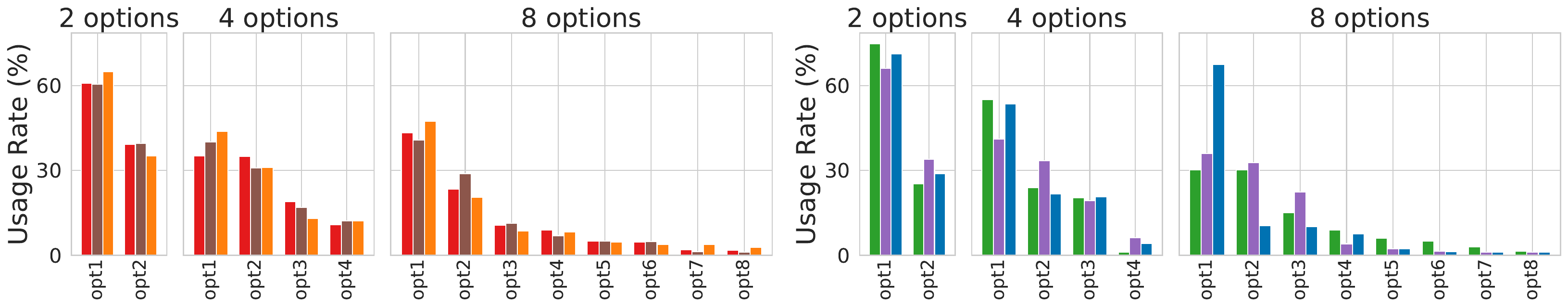}
    
    \begin{minipage}{0.95\linewidth}
        \centering
        \begin{tabular}{p{0.49\linewidth} p{0.49\linewidth}}
            \centering \small(a) FetchPush IOC options &
            \centering \small(b) FetchPush MOC options
        \end{tabular}
    \end{minipage}
    
    \vspace{0em}
    \begin{minipage}{0.95\linewidth}
        \centering
        \begin{tabular}{p{0.49\linewidth} p{0.49\linewidth}}
            \centering
            \small
            \scalebox{0.8}{
                \textcolor{ioc}{\rule[0.5ex]{2em}{1.5pt}} IOC \quad
                \textcolor{iocher}{\rule[0.5ex]{2em}{1.5pt}} IOC-HER \quad
                \textcolor{ioc2her}{\rule[0.5ex]{2em}{1.5pt}} IOC-2HER
            }
            &
            \centering
            \small
            \scalebox{0.8}{
                \textcolor{moc}{\rule[0.5ex]{2em}{1.5pt}} MOC \quad
                \textcolor{mocher}{\rule[0.5ex]{2em}{1.5pt}} MOC-HER \quad
                \textcolor{moc2her}{\rule[0.5ex]{2em}{1.5pt}} MOC-2HER
            }
        \end{tabular}
    \end{minipage}
    
    \caption{Option utilization rates for IOC (left panel) and MOC (right panel) methods with 2, 4, and 8 options in the FetchPush task. Bars show the mean usage of each option across 5 seeds after $1.5\times10^{3}$ iterations.}
    \label{fig:push_opts}
\end{figure}

\begin{figure}[t]
    \centering
    \setlength{\abovecaptionskip}{0pt} 
    \setlength{\belowcaptionskip}{0pt} 
    \includegraphics[width=\linewidth]{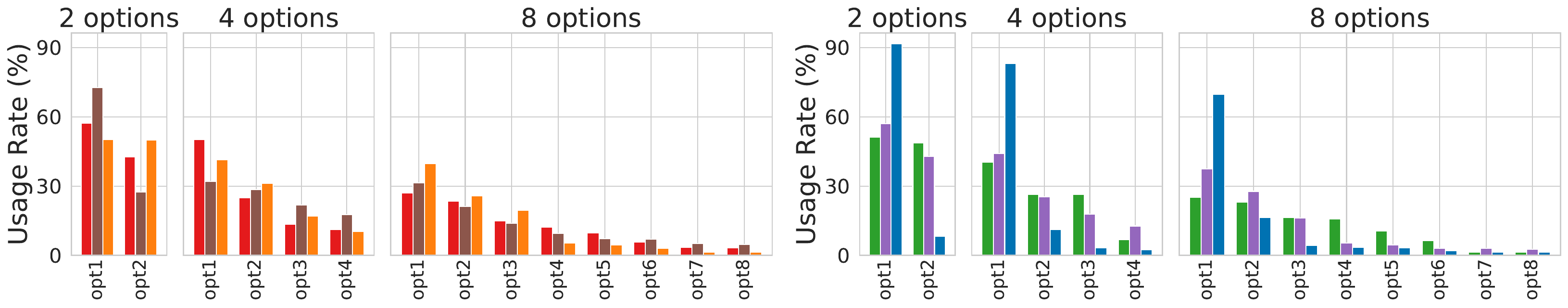}
    
    \begin{minipage}{0.95\linewidth}
        \centering
        \begin{tabular}{p{0.49\linewidth} p{0.49\linewidth}}
            \centering \small(a) FetchSlide IOC options &
            \centering \small(b) FetchSlide MOC options
        \end{tabular}
    \end{minipage}   
    
    \begin{minipage}{0.95\linewidth}
        \centering
        \begin{tabular}{p{0.49\linewidth} p{0.49\linewidth}}
            \centering
            \small
            \scalebox{0.8}{
                \textcolor{ioc}{\rule[0.5ex]{2em}{1.5pt}} IOC \quad
                \textcolor{iocher}{\rule[0.5ex]{2em}{1.5pt}} IOC-HER \quad
                \textcolor{ioc2her}{\rule[0.5ex]{2em}{1.5pt}} IOC-2HER
            }
            &
            \centering
            \small
            \scalebox{0.8}{
                \textcolor{moc}{\rule[0.5ex]{2em}{1.5pt}} MOC \quad
                \textcolor{mocher}{\rule[0.5ex]{2em}{1.5pt}} MOC-HER \quad
                \textcolor{moc2her}{\rule[0.5ex]{2em}{1.5pt}} MOC-2HER
            }
        \end{tabular}
    \end{minipage}
    
    \caption{Option utilization rates for IOC (left panel) and MOC (right panel) methods with 2, 4, and 8 options in the FetchSlide task. Bars show the mean usage of each option across 5 seeds after $1.5\times10^{3}$ iterations.}
    \label{fig:slide_opts}
\end{figure}

\begin{figure}[t]
    \centering
    \setlength{\abovecaptionskip}{0pt} 
    \setlength{\belowcaptionskip}{0pt} 
    \includegraphics[width=\linewidth]{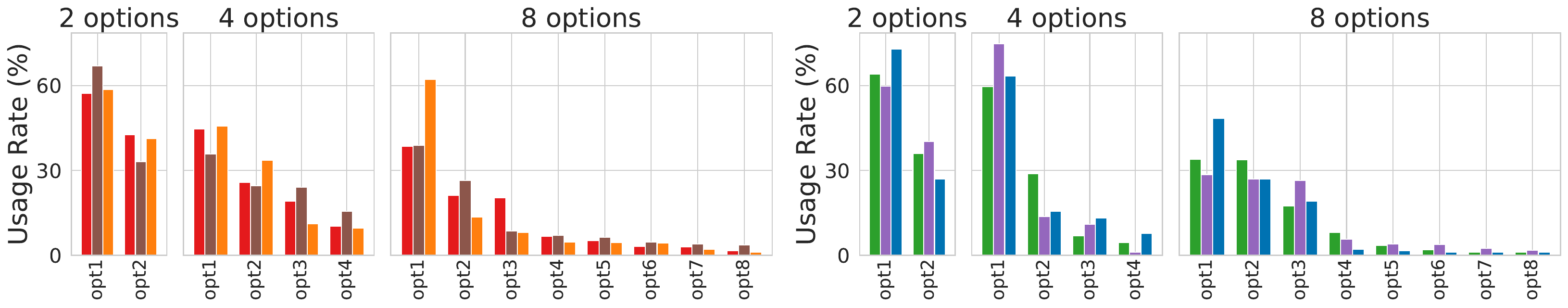}
    
    \begin{minipage}{0.95\linewidth}
        \centering
        \begin{tabular}{p{0.49\linewidth} p{0.49\linewidth}}
            \centering \small(a) FetchPickAndPlace IOC options &
            \centering \small(b) FetchPickAndPlace MOC options
        \end{tabular}
    \end{minipage}
    
    \vspace{0em}
    \begin{minipage}{0.95\linewidth}
        \centering
        \begin{tabular}{p{0.49\linewidth} p{0.49\linewidth}}
            \centering
            \small
            \scalebox{0.8}{
                \textcolor{ioc}{\rule[0.5ex]{2em}{1.5pt}} IOC \quad
                \textcolor{iocher}{\rule[0.5ex]{2em}{1.5pt}} IOC-HER \quad
                \textcolor{ioc2her}{\rule[0.5ex]{2em}{1.5pt}} IOC-2HER
            }
            &
            \centering
            \small
            \scalebox{0.8}{
                \textcolor{moc}{\rule[0.5ex]{2em}{1.5pt}} MOC \quad
                \textcolor{mocher}{\rule[0.5ex]{2em}{1.5pt}} MOC-HER \quad
                \textcolor{moc2her}{\rule[0.5ex]{2em}{1.5pt}} MOC-2HER
            }
        \end{tabular}
    \end{minipage}
    
    \caption{Option utilization rates for IOC (left panel) and MOC (right panel) methods with 2, 4, and 8 options in the FetchPickAndPlace task. Bars show the mean usage of each option across 5 seeds after $1.5\times10^{3}$ iterations.}
    \label{fig:pickandplace_opts}
\end{figure}

Figures \ref{fig:reach_opts}, \ref{fig:push_opts}, \ref{fig:slide_opts}, and \ref{fig:pickandplace_opts} further analyze option utilization, plotting the usage percentage of each option at the end of training for all scenarios. A key observation is that as the number of available options increases, agents tend to develop a clear preference for a smaller subset. However, in all successful agents leveraging 2HER or HER, no option was completely neglected; all options remained active in the agent's policy.

Conversely, for algorithms that failed to solve the task, option usage was typically more uniform. This is because the absence of successful trajectories prevents the credit assignment necessary for specialization; without positive reinforcement, no option learns a distinct, useful sub-policy, leading to near-random selection. Our experiments also highlight the difficulty of achieving balanced option usage in sparse reward settings. This reinforces the argument presented in the main paper that, for these scenarios, multi-option learning is more beneficial for improving performance and convergence stability through specialization than for maintaining a balanced utilization of all options.

\end{document}